\documentclass[letterpaper, 10 pt, conference]{ieeeconf}
\IEEEoverridecommandlockouts              
\usepackage{graphics}
\usepackage{epsfig} 
\usepackage{mathptmx}
\usepackage{times} 
\usepackage{amsmath}
\usepackage{amssymb}
\usepackage{cite}
\usepackage{here}
\usepackage{xcolor}
\usepackage{bm}
\usepackage{algorithm}
\usepackage{algorithmic}
\usepackage{array}
\usepackage{hyperref}
\usepackage{multirow}

\title{\LARGE \bf
 Robust Collision-free Lightweight Aerial Autonomy\\ for Unknown Area Exploration}
\author{Sunggoo Jung$^{1}$ Hanseob Lee$^{2}$, David Hyunchul Shim$^{2}$, and Ali-akbar Agha-mohammadi$^{3}$
\thanks{$^{1}$Sunggoo Jung is with the Autonomous UAV Research Section, ETRI, Daejeon, Korea. {\tt\small sunggoo@etri.re.kr}}
\thanks{$^{2}$Hanseob Lee, and David Hyunchul Shim are with the Unmanned Systems Research Group (USRG), KAIST, Daejeon, Korea. {\tt\small \{hslee89, hcshim\}@kaist.ac.kr}}%
\thanks{$^{3}$A. Agha-mohammadi is with NASA-JPL, California Institute of Technology, Pasadena, CA 91109 USA. {\tt\small aliagha@jpl.nasa.gov}}%
}
\begin{document}
\maketitle
\thispagestyle{empty}
\pagestyle{empty}
\begin{abstract}
Collision-free path planning is an essential requirement for autonomous exploration in unknown environments, especially when operating in confined spaces or near obstacles. This study presents an autonomous exploration technique using a small drone. A local end-point selection method is designed using LiDAR range measurement and then generates the path from the current position to the selected end-point. The generated path shows the consistent collision-free path in real-time by adopting the Euclidean signed distance field-based grid-search method. The simulation results consistently showed the safety, and reliability of the proposed path-planning method. Real-world experiments are conducted in three different mines, demonstrating successful autonomous exploration flight in environments with various structural conditions. The results showed the high capability of the proposed flight autonomy framework for lightweight aerial-robot systems. Besides, our drone performs an autonomous mission during our entry at the Tunnel Circuit competition (Phase 1) of the DARPA Subterranean Challenge.
\end{abstract}

%INTRODUCTION
\section{INTRODUCTION}
Autonomous exploration in an unknown cluttered environment poses a set of challenging problems in many levels\cite{balta2017integrated}, \cite{bircher2016three}. Unavailability of GPS signals and limited lighting conditions render the perception problem extremely difficult \cite{milford2012seqslam}. Confined spaces between walls and low height of priory unknown environment require compact drones which cannot rely on high-performance sensor-suites \cite{yap2009slam}. These constraints require careful attention in path-planning due to the existing state estimation errors. The path with hard constraints generates energy optimal shortest path to the goal. However, it may bring about too close path to walls or obstacles, making quadrotor flying unreliable when there are control or perception errors. Besides, it requires a large amount of computation as the environmental information grows. Thus, real-time behavior decisions of the drones are not easy to realize in large-area exploration applications.\\
To solve these problems, this study proposes a three-step path generation procedure to make the generated path strongly remain in the middle of the obstacles or walls to safely explore the unknown confined areas. A grid or graph search method requires the destination. In the case of unknown environment exploration, there are no goals or global plans. Thus, we use a frontier-based approach \cite{yamauchi1997frontier} and design a novel method to pick a receding-horizon local goal (we call it ``end-point'') using occupancy grid information. A high-level collision-free path is developed based on grid-search algorithm \cite{hart1968formal}. To make the path more reliable to fly, we generate the nearest occupied voxel information represented by the ESDF grid map and make the novel cost function of the grid-search method which penalizes the adjacent obstacle cost from ESDF information. Finally, we introduce a method to reduce the computational burden caused by the expansion of online maps as the exploration progresses. We employ the local-map approach which has 5m $\times$ 5m $\times$ 5m fixed map size to overcome this problem. The ESDF and local re-planning paths are generated within this local-map area to reduce the computational cost and achieve real-time performance. These aspects confer fully autonomous and robust exploration capability to real-world robots. This study has the following contributions:\\
\begin{figure}[!t]
    \centering
    \includegraphics[width=3.3in]{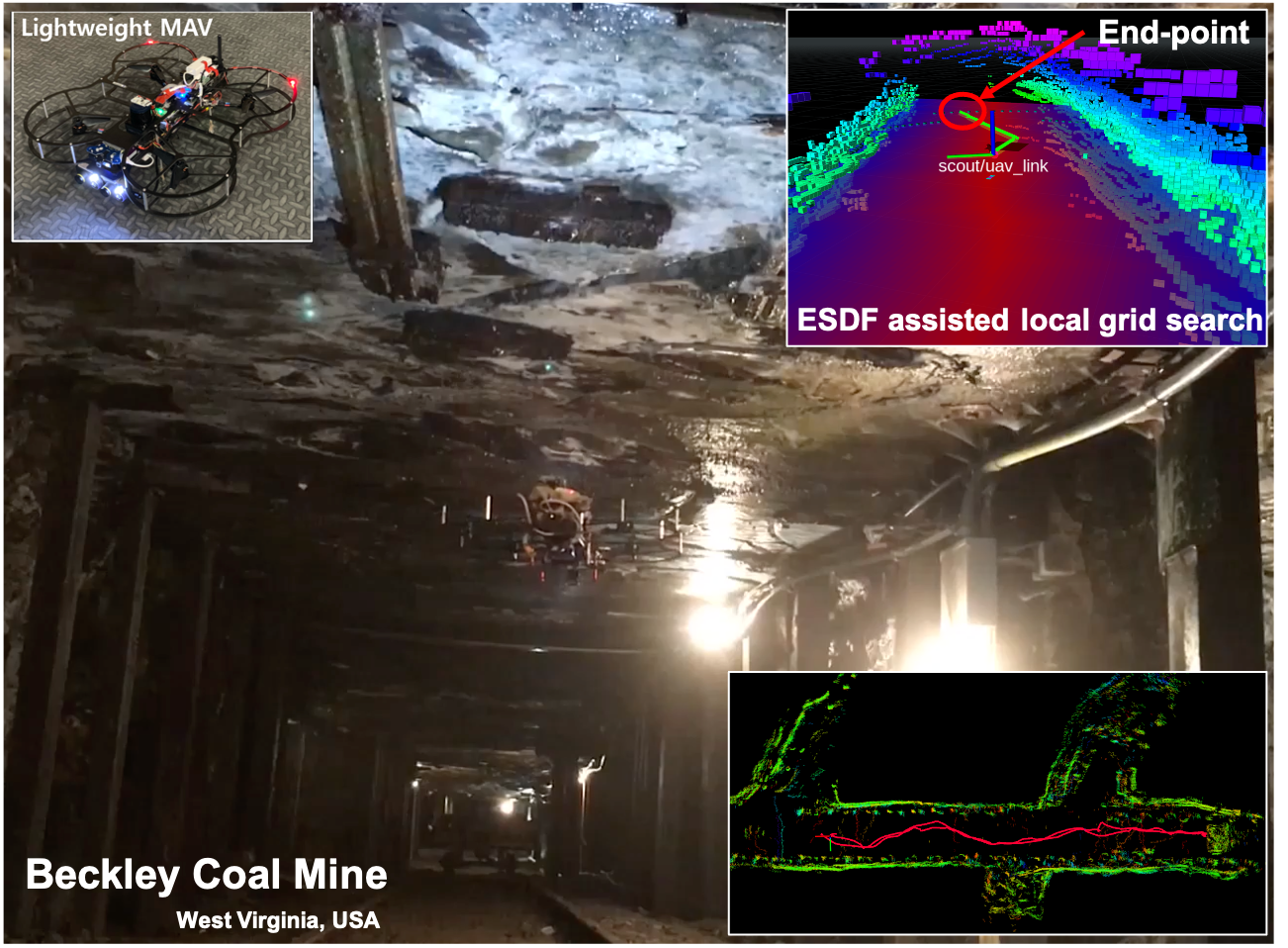}
    \caption{A snapshot of autonomous exploration experiment of the lightweight drone in coalmine environment\\ Location: Beckley mine, West Virginia, United States.}
    \label{fig:MAVSnap}
\end{figure}
$\bullet\ $ An autonomous aerial exploration implemented on small-sized (330 mm sized) flying robots with a fully on-board computation\\
$\bullet\ $ Robust and safely exploring the path that allows for real-time re-planning using local map approach\\
$\bullet\ $ Computational cost analysis by comparing state-of-the-art local path-planning methods\\
$\bullet\ $ Case study of a fully autonomous exploration, using the small aerial robot\\
$\bullet\ $ Public release of the code to the community. It can run on any type of (e.g., ground, aerial or underwater) robot for autonomous exploration applications.\\
\\
The remainder of the paper is organized as follows. In Section \ref{related_work}, we present an overview of related works. Section \ref{system_overview} describes the hardware and software of the proposed framework. In Section \ref{proposed}, we present the details of the proposed approach. The simulations and practical implementation results are illustrated and evaluated in Section \ref{E_R}. Section \ref{C_FW} presents the conclusions and discussions.

%RELATED WORKs
\section{RELATED WORK}
\label{related_work}
Considerable amount of methods have been investigated focusing on the problem of robotic exploration \cite{yamauchi1997frontier}, \cite{connolly1985determination}. The path planning uses mixed integer linear programming (MILP) \cite{kamal2005milp}, \cite{schouwenaars2004receding}. However MILP is an NP-hard; thus, it can only be used in a finite-size planning problem. MPC based receding horizon navigation methods \cite{kuwata2005robust}, \cite{shim2005autonomous} and urban environment exploration conceptual works \cite{prazenica2006vision}, \cite{bellingham2002receding} were proposed; however, they are limited by the finite area range.\\ Early results in unknown area exploration planner were proposed by \cite{bircher2016receding}. This study showed the capability of priory unknown area exploration while avoiding collisions. Kino-dynamic real-time planning using quadrotor has been presented in \cite{allen2016real}. This study showed full-stack planning architecture; however, the state estimation is rely on the VICON motion capture system. Front-end path planning and back-end path smoothing works have been proposed in \cite{gao2017gradient} and showed impressive results in a small range of indoor and outdoor areas. The MPC based online path planning work showed dynamic re-planning by considering multi-object concept\cite{peng2012intelligent}; however, the study finished within simulation without real-world validation. One-line segment combination path smoothing approach is presented in \cite{lai2016robust}. This study generateed the path using a two-point boundary value problem approach to reduce the computational cost. This study showed reliable flight results; however, the piece-wise segment of path joint optimization is not considered. Interesting works are presented by \cite{popovic2016online}. They developed an online path-planning algorithm to reduce energy costs while detecting weeds. They showed impressively reducing the energy cost by effective path-planning algorithm; however, only validated in simulation.\\
Recently, the large-area exploration problem has attracted significant attention because of DARPA Sub-Terranean Challenge \cite{darpa_rss}. Sampling-base method \cite{bircher2016receding}, graph-based approaches \cite{dang2020graph}, \cite{dang2018visual} and learning-based approaches have been proposed \cite{reinhart2020learning}. The above systems require high size, weight and power (SWAP) systems making the robot relatively heavy and large. Thus, the development of a compact-sized exploration framework is an essential challenge to explore confined and cluttered spaces.\\
Some solutions to the above have been presented such as resilient micro flyer design \cite{de2020collision} or range based sensor array approach \cite{tan2019design}. These demonstrate simply penetrating a small-sized tunnel or structured area, which is impractical for large-area exploration.\\
Motivated by these studies, we discovered that the study for ``\textit{lightweight autonomy}'' in confined and cluttered areas has not been properly investigated. This contribution aims to provide a meaningful low SWAP system alternative by combining both algorithms, the local end-point selection and the ESDF-assisted collision-free path, with fully on-board computing.

\section{SYSTEM OVERVIEW}
\label{system_overview} 
In this study, which in turn focuses on the confined space exploration, a ``350-size'' frame is designed for lightweight autonomy. The designed concept of the system is lightweight but high performance with an acceptable flight time. The platform has four DJI Snail racing drone propulsion systems, which can lift all the sensor systems required for autonomous exploration in 12 min. We chose the PixRacer R15 autopilot with PX4 Stack for low-level flight control and NVIDIA Jetson TX2 GPU for high-level mission management. The TX2 module is mounted on a third-party carrier board (Auvidea, J120 \cite{AuvideaJ120}), which has RS-232 serial ports for communication with the PX4 flight controller.\\
With the constraint of the payload limit, the drone is equipped with a two-dimensional (2D) laser scanner (Hokuyo, UST-20LX) instead of three-dimensional (3D) LiDAR, such as the popular Velodyne VLP-16. It is worth noting that in the texture-less environment, the localization using only the laser scanner can be ineffective since the features on the walls are monotonous. Thus we integrate an upward-facing Qualcomm Snapdragon Flight (QSF), which delivers on-board visual-inertial-odometry and its fisheye camera data for localization robustness. For altitude measurement, time-of-flight based 1D LiDAR is equipped (Terabee, TeraRanger-Evo). To aid and offer robust perception to vision source in the dark environment, we integrate an upward-facing high-power LED (Lume, Cube 2.0, housing case uncovered). The above sensor data are run on the NVIDIA Jetson TX2 (Ubuntu 18.04 variant and ROS Melodic). The total weight of the system, including all the sensing and processing components, and the battery is 1.43 kg.

\section{Proposed Approach}
\label{proposed}
\begin{figure}[!t]
    \centering
    \includegraphics[width=3.3in]{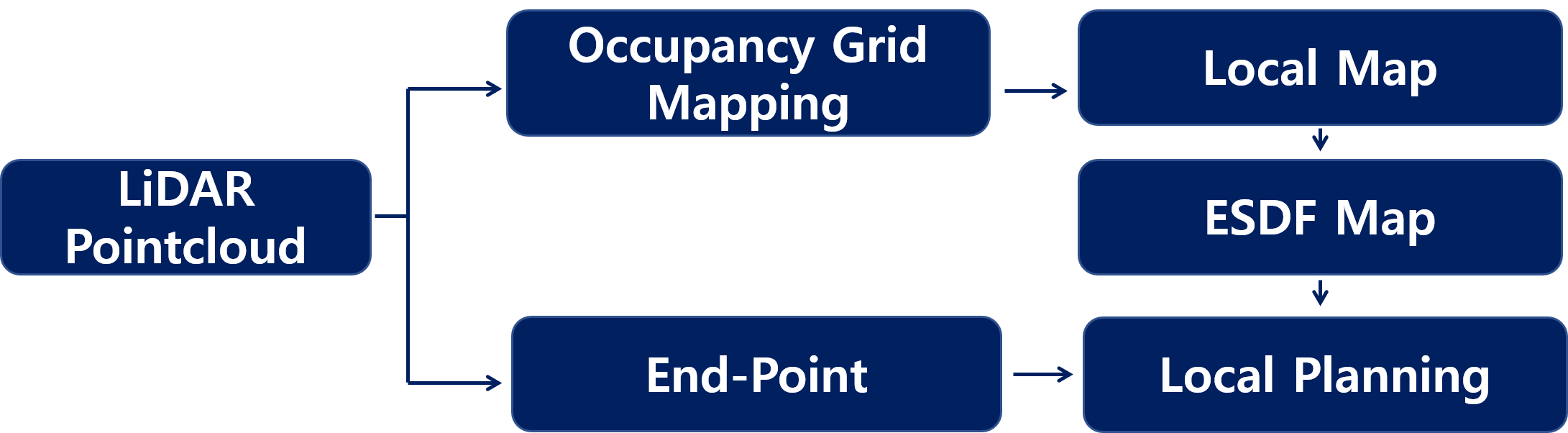}
    \caption{Diagram of lightweight autonomy framework. An ESDF map is developed using the mapped knowledge. ESDF-assisted planner determines good exploration paths starting at the current location to the end-point.}
    \label{fig:software}
\end{figure}

The proposed lightweight flight autonomy method consists of three collaborative modules: ESDF-assisted path planning, local-map generation, and end-point selection. The structure of this collaborative module setup, is shown in Figure \ref{fig:software}. First, the proposed method generates occupancy representation \cite{45466} and calculates ESDF, which allows collision-free paths. The planning module exploits the ESDF cost using the grid-search method, offering obstacle-aware safety paths. Second, to reduce the computational cost and ensure real-time performance for computing cost-constrained lightweight drones, every iteration of the path-planning generates the trajectory within 5 m $\times$ 5 m $\times$ 5 m local-map around the current robot location. Finally, the proposed method introduces the end-point selection to make the planner generates trajectory within the local-map. As the exploration progresses, the occupancy information around the robot location changes. If the end-point, the destination of the path planner, is at the obstacle, the correct path cannot be created. Thus, the occupancy data and pointcloud sensor information is used to select the obstacle-free end-point. This section discusses these methods with respect to the lightweight flight autonomy concept.
\subsection{Algorithm Description}
Algorithm \ref{alg:ESDF} outlines path-planning steps of the proposed framework. Every iteration starts by acquiring an occupancy probability map (${M_{prob}(\cdot)}$) from the SLAM module. From the certain threshold condition ($T_{\zeta}$), ${M_{prob}(\cdot)}$ is converted to Boolean-valued voxel grid, marking flyable area by value of one (true) and fill the rest area with zero (false) ($M_{bool}(\cdot)$). $M_{bool}$ is used to generate an ESDF-map representing distance information from the adjacent obstacle. The local end-point selection module receives current state ($\xi = [x, y, z, \psi]$),  LiDAR sensor data ($\sigma_k$). This module find best end-point $b_{k}^{*}$ from the end-point candidate $S_k$. Then, the $b_{k}^{*}$ passes a low-pass filter to prevent oscillation of the selected point. Finally, the planner receives the current state, selected end-point and ESDF info to generate a collision-free trajectory. It uses cost modified ESDF-assisted grid search. The key functionalities are explained in detail in the following subsections.
\begin{algorithm}
\caption{ESDF-assisted Path} \label{alg:ESDF}
\begin{algorithmic}[1]
\STATE \textbf{function} genESDF$(M_{prob}(\boldsymbol{p}))$
\STATE \quad\textbf{for} each grid point $\boldsymbol{p}$, \textbf{in} $M_{prob}(\boldsymbol{p})$
\STATE \qquad\textbf{if} $M_{prob}(\boldsymbol{p}) > T_{\zeta}$
\STATE \quad\qquad$M_{bool}(\boldsymbol{p}) \leftarrow \text{true}$
\STATE \textbf{function} \text{ESDT($M_{bool}(\cdot)$)} \textbf{return} $ESDF$
\STATE \textbf{function} \text{FindEndPoint($\xi, \sigma_k, \psi_k$)}
\STATE \quad\textbf{if} $\sigma_k \geq inf$
\STATE \qquad$S_k \leftarrow \text{5m}$
\STATE \quad\textbf{else if} $4m \leq \sigma_k \leq 15m$
\STATE \qquad$S_k \leftarrow \text{4m}$
\STATE \quad\textbf{else} $S_k \leftarrow \sigma_k$
\STATE \quad\text{findBkStar($S_k, \psi_k$)}
\STATE \quad\text{lowPassFilter($b_{k}^{*}, freq$)}
\STATE \textbf{function} \text{Planner($\xi, b_{k_{lpf}}^{*}, ESDF$)}
\STATE \quad\textbf{function} \text{InitializePathFinder($\xi, b_{k_{lpf}}^{*}$)}
\STATE \quad\textbf{function} \text{FindPath($\xi, b_{k_{lpf}}^{*}, ESDF$)}
\STATE \qquad\text{convert pose $\xi$ to grid node $n$}
\STATE \qquad\text{find a path using grid search}
\STATE \qquad\text{where, $f_{total}(n) = f_{score}(n) + f_{dist}(n)$}
\STATE \quad\textbf{return} path
\end{algorithmic}
\end{algorithm}

\subsection{Local-map approach}
The computational complexity of the grid or graph search-based planner depends on the map resolution and size. In the fixed-horizon path-planning problem, the user usually sets the admittable map size priory to ensure the planner performance. The unknown area exploration problem, the area to search, is not known. Thus, as the exploration goes, the computational complexity grows in a cube of grid size ($O(n^3),\, n=grid\, size$), which is impractical for a low-cost embedded on-board processor. The proposed local-map approach is designed to reduce the computational cost by performing local planning in local-map area $[sL, sW, sH]$ and resolution $r$. Thus, the time complexity remains fixed $O(sLsWsH/r^3)$ since the local-map maintains a constant size regardless of the exploration progress. The complexity only depends on the volume of the local-map when the map resolution is fixed. The local-map is updated at every 0.1 s (10 Hz); thus, this method can assure the vehicle velocity of about 50 m/s.

\subsection{End-point Selection}
\begin{figure}[!t]
    \centering
    \includegraphics[width=3.3in]{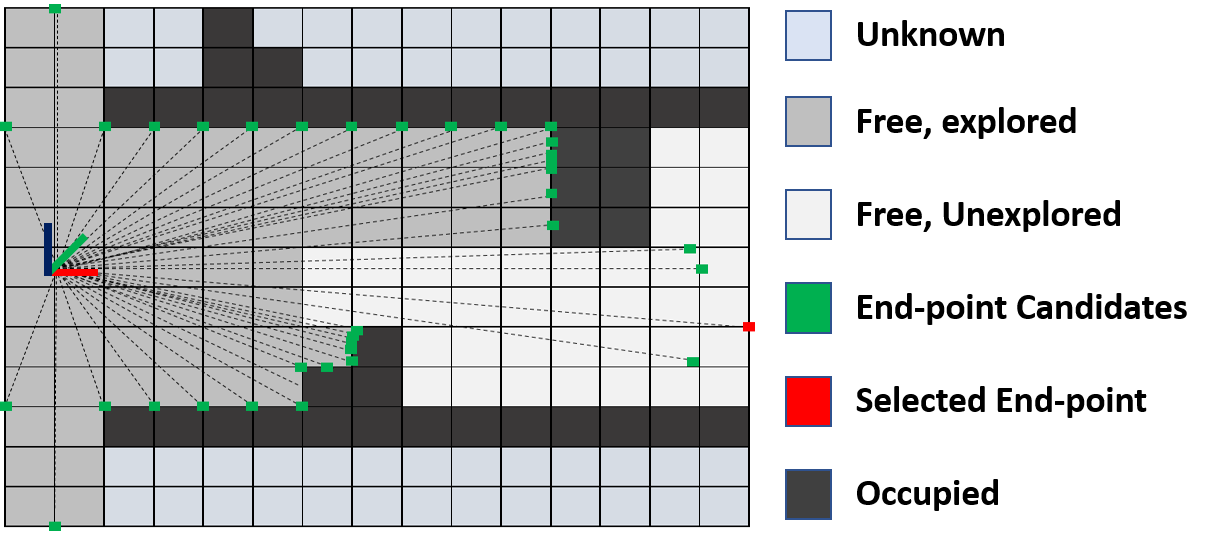}
    \caption{Example illustration of the end-point selection using \autoref{eq:thres}, \ref{eq:goal}}
    \label{fig:endpoint}
\end{figure}

As discussed above, when exploring areas fully unknown \textit{a priori}, to use search-based planning method, the destination must be defined in advance. Therefore, the ``end-point,'' the destination in local-map area must be determined based on the obstacle information from local-map window. To resolve this problem, we implement the end-point selection method using the range measurements for search-based planning in the unknown area. The range sensor gives the index and distance of the point denoted as $[k, \sigma_k]$. To select the end-point, we sample evenly 48 points out of 1440 point clouds from the laser scanner range sensor. The samples are at 30 intervals starting from the 1st index of the points among a total of 1440 raw points of laser scanner inputs(see, Fig. \ref{fig:endpoint}). This constitutes a set of 48 sample points surrounding the vehicle at 240 degrees (the sample field of view $\Omega$). Next, we collect a set of points exceeding a certain distance $l$. Here, ``$l$'' is the design factor, maening the planning range of the local planner. For instance, if $l$ is set to 5 m, the local planner plans the path 5 m from current vehicle pose $\xi$. Then, the method determines the maximum index of this set and converts the pose of its points into a body frame coordinate system. Finally, the point is selected as the end-point (Eq. \ref{eq:goal}.)

\begin{equation} 
\label{eq:thres}
S_k = \left\{\begin{matrix} l & if\;\; \sigma_k = \infty,\:\;\;\;\;\;\;\;\;\;\;\;\;\;\;\\[2mm]
l-1 & if\;\; l-1 \leq\sigma_k\leq l_{max},\\[2mm]
\sigma_k & otherwise,\;\;\;\;\;\;\;\;\;\;\;\;\;\;\;\;\;\;
\end{matrix}\right.
\end{equation}

\begin{equation}
\label{eq:goal}
S^{*}=\underset{k\in N}{max}\;\;\xi_k + S_k\cdot \binom{cos(\psi_k+k\cdot\frac{\Omega}{N}-\frac{\Omega}{2})}{sin(\psi_k+k\cdot\frac{\Omega}{N}-\frac{\Omega}{2})}
\end{equation}

where the sampled points $S=\{S_k\}_{k=1,...,K},S_k\in \mathbb{R}{^{2}}$ and total number of samples $N$. Figure \ref{fig:endpoint} shows the end-point selection example in three different path shapes, such as straight, right open space, and four-way junction. The selected $S^{*}$ is defined as the end-point of the grid-search path. Finally, a low-pass filter is implemented to prevent end-point oscillation due to the noise of raw laser scanner input.

\subsection{ESDF Weighted Cost Function} 
The online map from the cartographer (\cite{45466}) gives the map of the environment in a regular probability grid form. With this probability, we generate a Boolean-valued voxel representation which characterizes the flyable area, defined as
\begin{equation} 
\label{occupancy_prob} 
M_{bool}(\xi) = \left\{\begin{matrix} 1 & if\;\; M_{prob}(x)\: > T_{\zeta},\:\: \\[-4mm]
\\0 & otherwise,\;\;\;\;\;\;\;\;\;\;\;\;\;\;\;\:\:\:
\end{matrix}\right.
\end{equation}
where ${M_{prob}(\cdot)} : \mathbb{R}{^{2}}\rightarrow \mathbb{R}$ is the occupancy probability function of the local-map at grid point $\boldsymbol{x}$, ${M_{bool}(\cdot)}$ is the Boolean-value representation, and $T_{\zeta}$ is the constant probability threshold value (which is set for 70 (range: 0-100) in this work). ${M_{bool}(\cdot)}$ is then converted to distance field using Euclidean signed distance transform \cite{oleynikova2016signed}.\\
For planning, we implement a grid-search method to determine the shortest path from the vehicle`s current position to the selected end-point. We model the cost function using the following two terms: an ESDF term $f_{dist}$, which measures the cost of being near obstacles; and a prior term $f_{score}$, which measures the distance of the path between the current position and end-point. Therefore, we present the objective as follows:
\begin{equation} 
\label{eq:cost}
f_{total}(n) = f_{score}(n) + f_{dist}(n)
\end{equation}
\begin{equation}
f_{score}(n) = g(n) + (1+\epsilon\cdot\omega(n))h(n)
\end{equation}
\begin{equation} 
\omega(n) = \left\{\begin{matrix} 1-\frac{\eta(n)}{N} & if\;\; \eta(n)\: \leq N\\[-4mm]
\\0 & otherwise\;\;\;\;\;\;\;\;
\end{matrix}\right.
\end{equation}
where $g(n)$ is the cost from start to node $n$, $h(n)$ is the cost from $n$ to goal, $\eta(n)$ is the depth of the search and $N$ is the estimated length of the path to reach the goal \cite{pohl1973avoidance}.
\begin{figure}[t]
    \centering
    \includegraphics[width=3.3in]{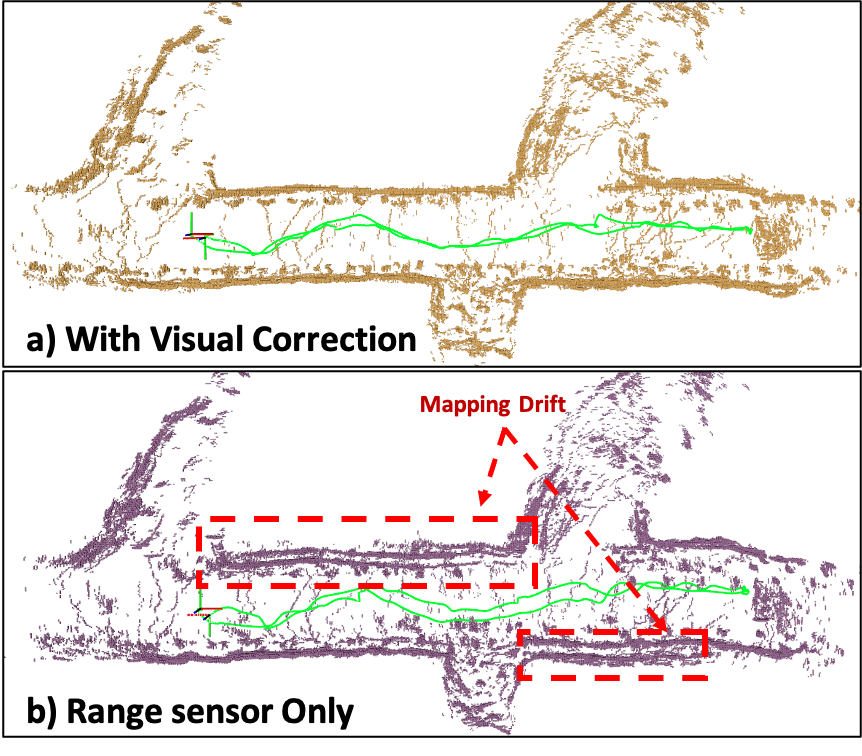}
    \caption{Experimental comparison of localization with a visual correction (top) and LiDAR-only localization (bottom). LiDAR-only SLAM shows mapping drift of 0.14/m due to the lack of features of underground mine environment. \\Location: Beckley Mine in West Virginia, United States}
    \label{fig:vioCompare}
\end{figure}
\subsection{Localization \& Mapping}
Underlying mapping representation of this study takes Cartographer by Google \cite{45466}, which showed an outstanding performance using continuous-time SLAM by Ceres-based scan matcher and optimizer \cite{ceres-solver} and grouped probability grids called submaps. The Cartographer receives an odometry source as a secondary input to correct position drift in a feature-less environment. Therefore, we employ a vision-based state-estimation source from QSF as an additional measurement source. Figure \ref{fig:vioCompare} shows the mapping result of LiDAR-only SLAM and SLAM with a visual correction at the Beckley mine in West Virginia. Since we fuse vision-based state estimation from QSF board, which are highly integrated single-board computers, time synchronization is performed with Chrony \cite{chrony} to cooperate this board with mission computer.
\begin{figure}[H]
    \centering
    \includegraphics[width=3.3in]{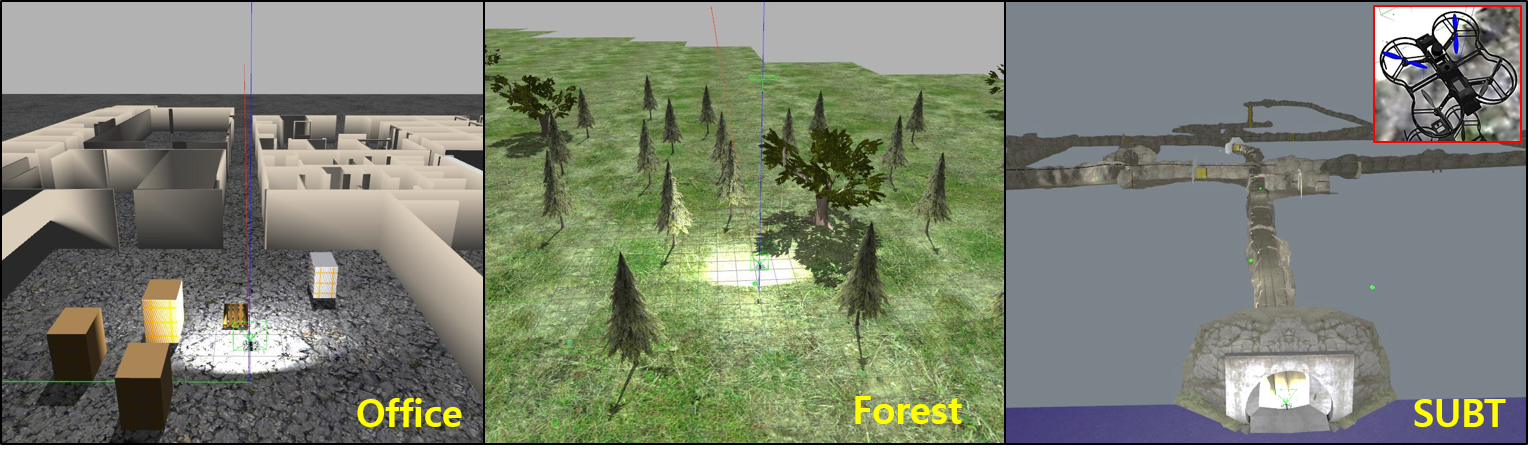}
    \caption{Various simulation environments setup with custom build 350 mm lightweight drone to evaluate and fine-tune the proposed method.}
    \label{fig:simulation}
\end{figure}
\begin{figure}[!t]
    \centering
    \includegraphics[width=3.4in]{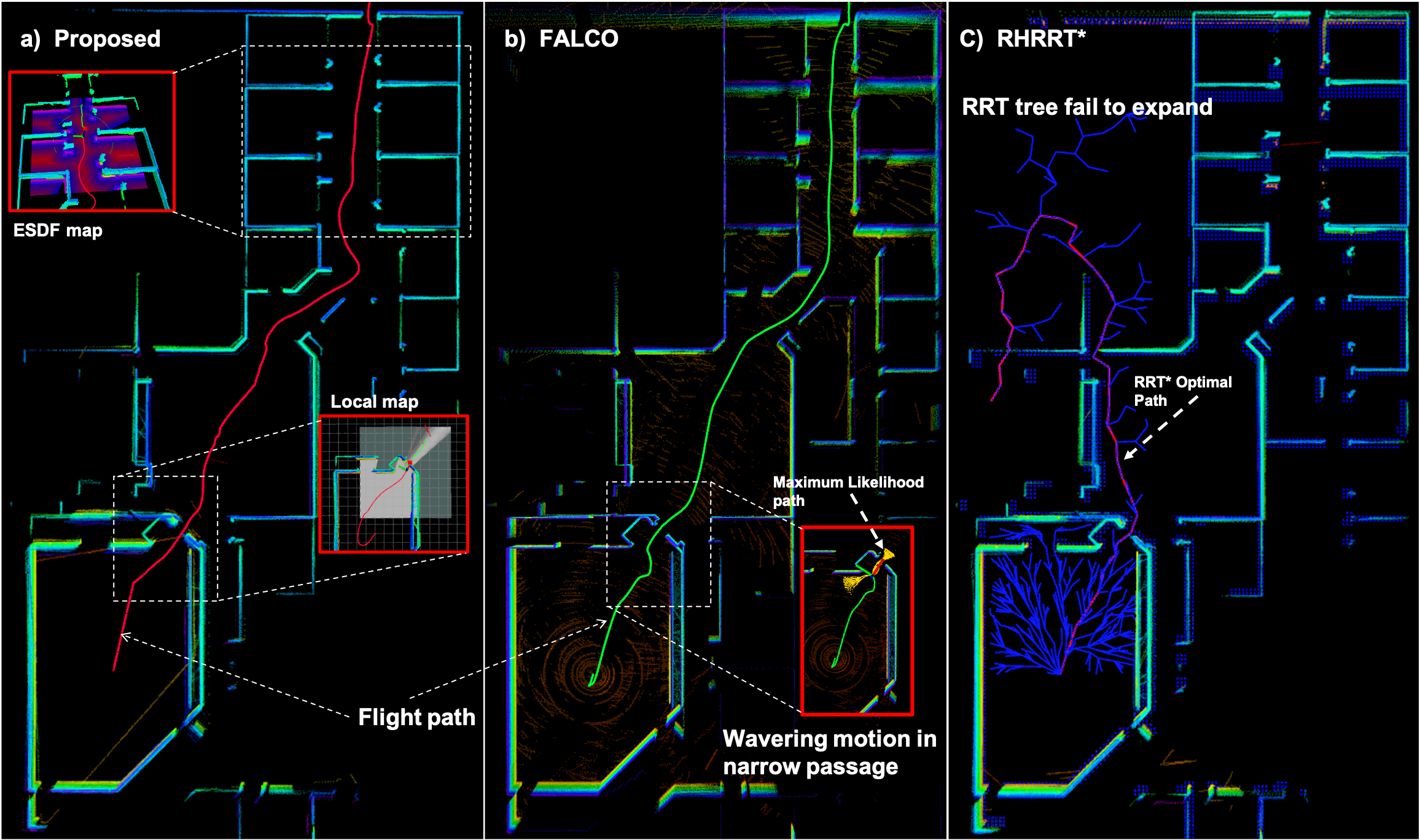}
    \caption{Compared confined office area simulation results of flight autonomy frameworks in the local planning viewpoint. The proposed lightweight flight autonomy and Falco show good performance, but RHRRT$^*$ cannot determine the proper path. The proposed lightweight framework, as shown in (a), successfully finds the safe path. The snapshots of the local-map and ESDF map at each position are presented. The solutions from Falco show the overall good performance; however, it presents wavering or staying motion for a few seconds since it gets stuck in local-minima, as shown in (b). The RHRRT$^*$ could not determine the path until the spacious amount of time since it spans the trees over the entire map.}
    \label{fig:simComp}
\end{figure}
\section{Evaluation Studies}
\label{E_R}
To evaluate the lightweight autonomy framework, a series of simulations and experimental studies are conducted. Within the experimental field results, experiments in three different underground mine environments were performed to validate the robustness and scalability of the system. The experiment environments include the coal-mine in South Korea and gold and coal mines in the United States.

\subsection{Simulation Results}
To evaluate and compare the proposed lightweight flight autonomy with other state-of-the-are method, the simulation study is conducted before its real-world verification. Simulations can perform large-scale iterative tests without any cost loss. This allows us to test the stability and capability of the newly applied algorithms before conducting hardware experiments. Three simulation configurations consisting of an office, forest, and subterranean environments were utilized \ref{fig:simulation}. The office environment consists of narrow hallway and multiple rooms. The forest environment consists of cluttered trees and unstructured models. Finally, the subterranean environment consists of multi-way intersections and narrow corridors, which are multiple kilometers long. Examples are performed using a high-fidelity Gazebo simulator with a Px4 flight stack \cite{meier2011pixhawk} for flight performance evaluation. 

\begin{figure}[!t]
    \centering
    \includegraphics[width=3.3in]{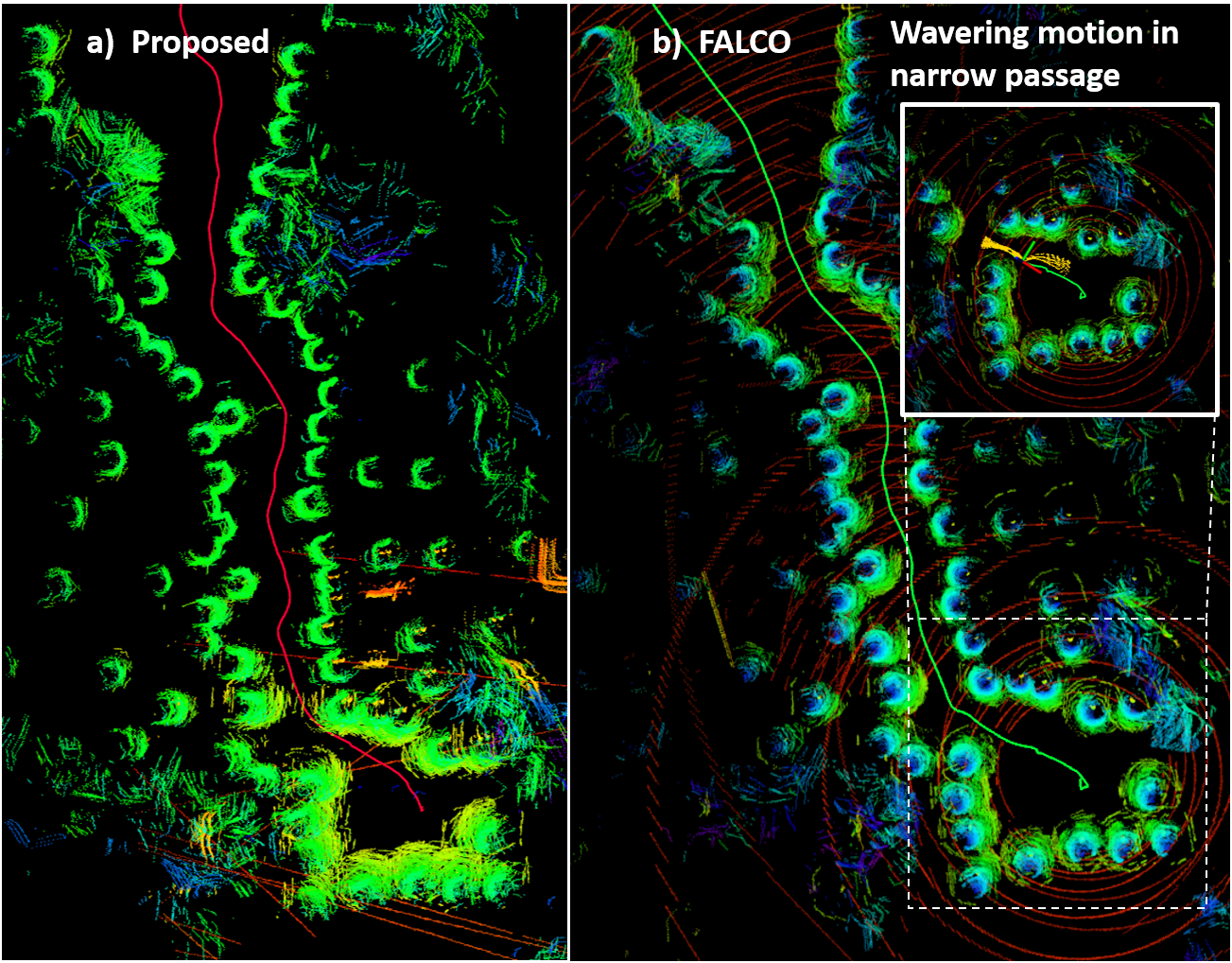}
    \caption{Compared forest area simulation results of flight autonomy frameworks in the local planning viewpoints. The proposed lightweight flight autonomy and Falco show good performance; however, the Falco shows wavering or staying motion in a narrow passage.}
    \label{fig:simCompForest}
\end{figure}
\begin{figure}[!t]
    \centering
    \includegraphics[width=3.4in]{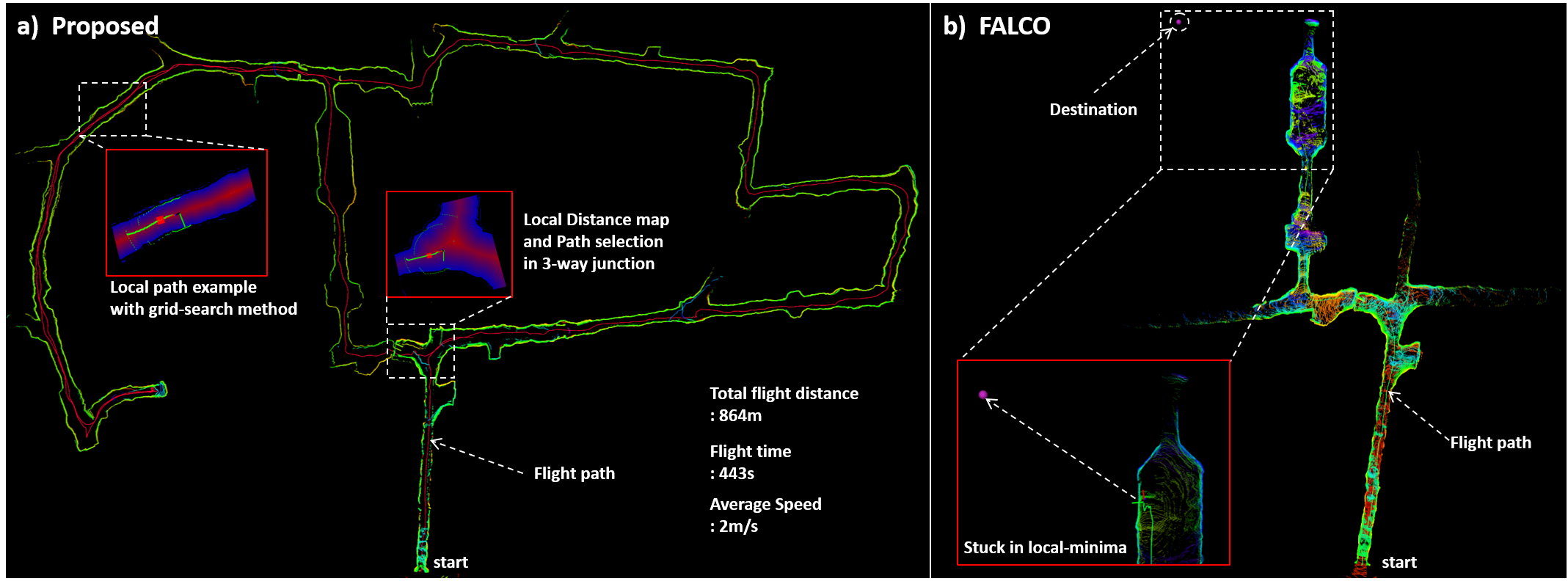}
    \caption{Compared underground mine (subterranean) area simulation results of flight autonomy frameworks in the local planning point of view. Our planner shows best performance while flying over 864m in 443s as shown in (a). Falco shows reasonable performance, however, it stuck in local-minima and cannot proceed further as shown in (b).} 
    \label{fig:simCompSubt}
\end{figure}
\begin{figure}[!t]
    \centering
    \includegraphics[width=3.4in]{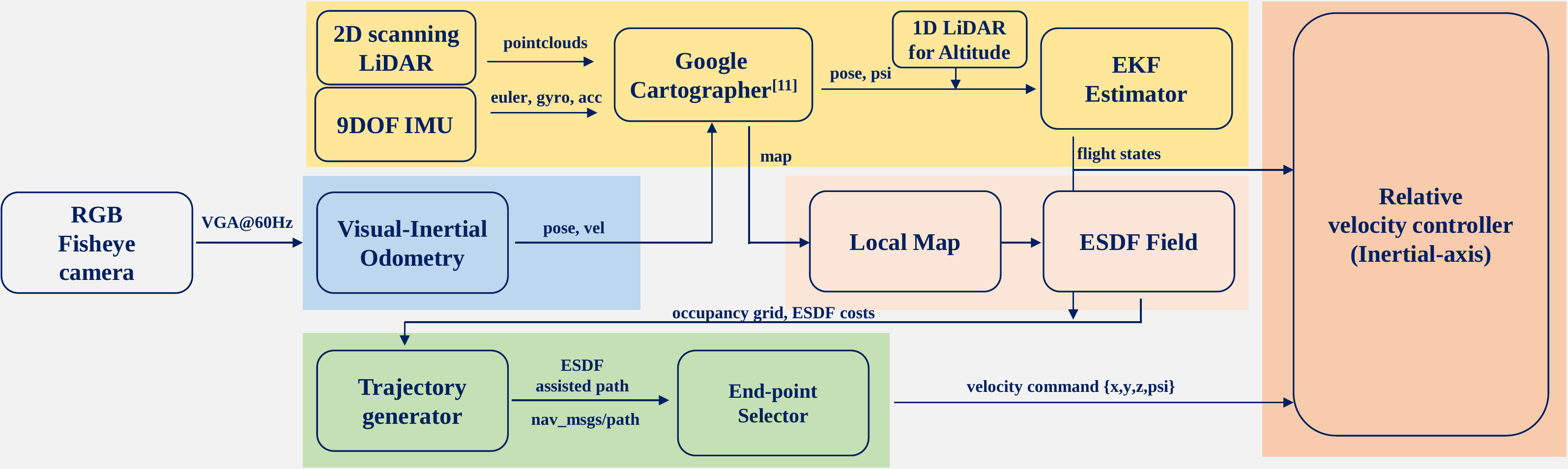}
    \caption{Software architecture overview of our lightweight autonomy system from sensor input to control command output.}
    \label{fig:ex_architec}
\end{figure}
The custom robot is developed for simulation studies assuming a quadrotor 350 mm MAV model with a laser scanner, a 2D rotating scanner up to 20 m. The local planning window is set to 5 m $\times$ 5 m $\times$ 5 m around the current vehicle pose, and the flight speed was set to 2 m/s. In all simulated tests the performance comparison is conducted with receding horizon RRT$^*$ (RHRRT$^*$) \cite{karaman2011anytime} and likelihood-based planning \cite{zhang2020falco}. Simulation results presenting the performance of the (a) proposed planner, (b) Falco, and (c) $RHRRT^*$ planner are shown in Figures \ref{fig:simComp}, \ref{fig:simCompForest} and \ref{fig:simCompSubt}. In these simulation tests, all methods use the same vehicle configuration, except for the range measurement sensor. Our planner and RHRRT$^*$ used 2D laser scanner-based 2.5D mapping while Falco used LOAM \cite{zhang2014loam} based 3D mapping method due to the basic framework setting difference. Our method and Falco show the best performance. The proposed lightweight framework successfully explores the three simulation environments. Falco shows reasonable performance. However, it shows staying or back-and-forth motion since to it cannot define the next step that maximizes the likelihood for the drone to fly to the goal. Besides, Falco needs destination points. Thus, its approach inadequate for the unknown area exploration problem, which cannot define the destination in advance at least in the current Falco open-source package. The RHRRT$^*$ method was excluded from the forest and subterranean simulation test because consumes much time to determine. Thus, it mostly fails to obtain a meaningful result.
Comparative analysis from simulation studies showed that the proposed lightweight autonomy framework has appropriate capability in various environments (see \autoref{table:comp}). In three scenarios, the exploration time for each method was tested five times in the same initial configuration. In all tests, the maximum flight speed was set to 2 m/s. Falco and the proposed planner have successfully reached the destination in office and forest tests. On average, the proposed planner performs more than 10 s faster, as Falco showed a time lag while scrambling in a narrow passage. For underground tests, Falco was unable to escape from the forked road where one road was blocked. This is considered an inevitable problem because Falco proceeds by changing only the heading while looking at only the small local area. An extensive set of field experiments is presented in the next subsection to further verify of the proposed framework in a real-world environment.
\begin{table}[!t]
\centering
\caption{Performance comparison in each scenarios}
\label{table:comp}
\begin{tabular}{lllllllllll}
\hline
\textbf{Scene} & \textbf{Method} & \multicolumn{4}{c}{\textit{\textbf{Exploration Time(s)}}} \\ \cline{3-6}
& & \textbf{Avg} & \textbf{Std} & \textbf{Max} & \textbf{Min} \\ \hline 
\multirow{3}{*}{\textbf{Office}} & Falco \cite{zhang2020falco}& 67.17& 6.0& 72.05& 58.36 \\
   & RHRRT$^*$& -& -& -& - \\
   & Proposed& 40.89& 1.37& 42.48& 39.17 \\ \hline
\multirow{2}{*}{\textbf{Forest}} & Falco \cite{zhang2020falco}& 59.83& 7.74& 69.29& 49.89 \\ 
   & Proposed& 42.19& 1.49& 43.78& 40.11 \\ \hline
\multirow{2}{*}{\textbf{Subt}} & Falco \cite{zhang2020falco}& -& -& -& - \\  
   & Proposed& 226.00& 1.13& 227.43& 224.56 \\ \hline
\end{tabular}  
\end{table}

\subsection{Experiments}
In this section, we evaluate the performance of the proposed autonomous subterranean exploration strategy in various real-world mine environments. The Eagle mine in the United States, California, has a long narrow straight passage. The Hawsoon mine in South Korea has a completely dark environment and round-shape structure. The Beckley mine in the United States, West Virginia, has a square shaped structure and many pillars inside the mine. All flight experiments were performed in fully autonomous mode in several test sites, which were unknown in advance. The overall system architecture is illustrated in \autoref{fig:ex_architec}.
\begin{table}[H]
\centering
\caption{Characteristic of Eagle Mine}
\label{table:goldmine}
\begin{tabular}{|c|c|c|c|}\hline
\textit{\textbf{Parameter}} & \textit{\textbf{Value}}& \textit{\textbf{Parameter}}&\textit{\textbf{Value}} \\ \hline
Width & $\leq 1.2 $ m& Name&Eagle mine \\ \hline 
Height & $\leq 1.0 $ m& Location & Southern California \\ \hline 
Tunnel Shape & $\cap$ shape  & Characteristic & Dusty, Narrow \\ \hline
\end{tabular}
\end{table}
\subsubsection{Test in the Eagle Mine}
\begin{figure}[!t]
    \centering
    \includegraphics[width=3.3in]{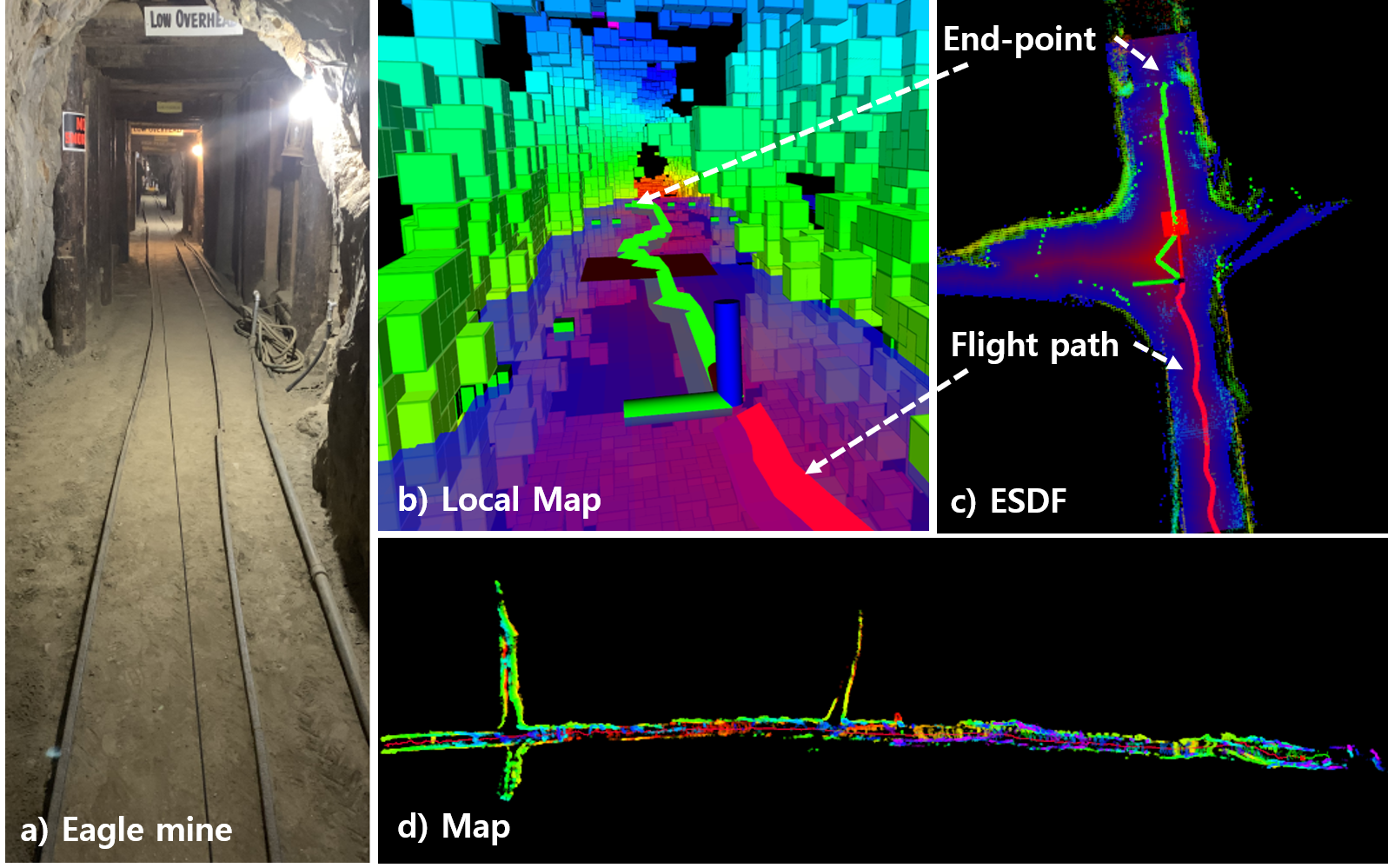}
    \caption{Experiment result at the narrow goldmine environment (a). b) Local-map of straight path, c) ESDF Map and consequently local-path planning results at four-way junction.\\Location: Eagle mine, United States, Southern California}
    \label{fig:eagleMine}
\end{figure}

In this example, the drone explores a long narrow wall shaped gold-mine, located in Southern California. Here, we evaluate the proposed algorithm integration performance. The end-point selection, local-map generation, ESDF in local-map area, and the grid-search algorithm. As presented in \autoref{fig:eagleMine}, the proposed lightweight autonomy framework is successfully integrated. The drone starts at the entrance of the mine and explores about 105 m in 62 s while passing a four-way junction. All environmental information is unknown before its flight test. The characteristics of this test site are presented in Table \ref{table:goldmine}.

\begin{figure}[!t]
    \centering
    \includegraphics[width=3.4in]{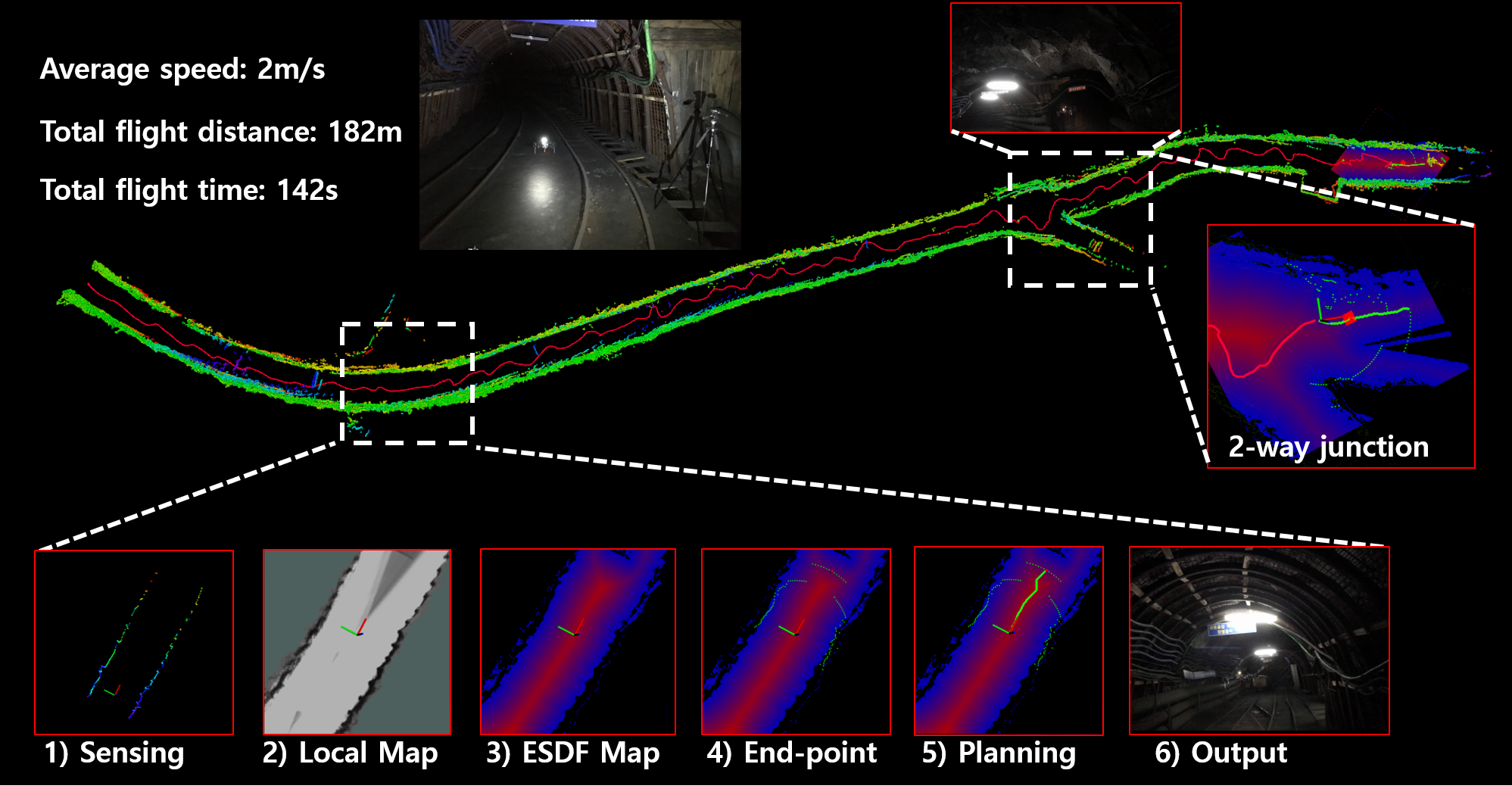}
    \caption{Autonomous exploration results at the Hwasoon coalmine in south Korea. An example of exploration step is presented with fully autonomous flight. The drone explores about 182 m during 142 s.\\ Location: Hwasoon mine (active mine), South Korea.}
    \label{fig:coalmineKR}
\end{figure}

\begin{table}[b]
\centering
\caption{Characteristic of Hwasoon Mine}
\label{table:coalmineKR}
\begin{tabular}{|c|c|c|c|}\hline
\textit{\textbf{Parameter}} & \textit{\textbf{Value}}& \textit{\textbf{Parameter}}&\textit{\textbf{Value}} \\ \hline
Width & $\leq 4.0$ m& Name&Hwasoon coal mine \\ \hline 
Height & $\leq 2.5$ m& Location & South Korea \\ \hline 
Tunnel Shape & $\cap$ shape   & Characteristic & Somewhat Dusty \\ \hline    
\end{tabular}
\end{table}
\subsubsection{Test in the Hwasoon Mine}
In this example, we present a coalmine exploration with a two-way junction and smooth curves terrain. Unlike the square-shaped narrow long-wall goldmine example presented in the previous section, the coalmine in South Korea has a round-shaped and relatively wide width. The coalmine in South Korea has an uneven and bumpy on the walls because the miners widen the mine with a shovel. Besides, wooden supports are supported every 1 m to prevent the mine from collapsing. Therefore, it is useful for scan-matching with enough features on the wall, but it is not easy to use for vision sources because of the dust on the floor. In this experiment, the quadrotor successfully explores 182 m for 142 s along a curved tunnel using the lightweight autonomy framework. Figure \ref{fig:coalmineKR} shows all the integrated modular framework. More specifically, at the two-way junction, the end-point selection method found that the right forked path was blocked and chose the left path. This test allowed us to evaluate long-range flight capabilities 
with dead-end intersections. The specification of this test mine is presented in Table \ref{table:coalmineKR}.

\subsubsection{Test in the Beckley Mine }
\begin{figure}[H]
    \centering
    \includegraphics[width=3.4in]{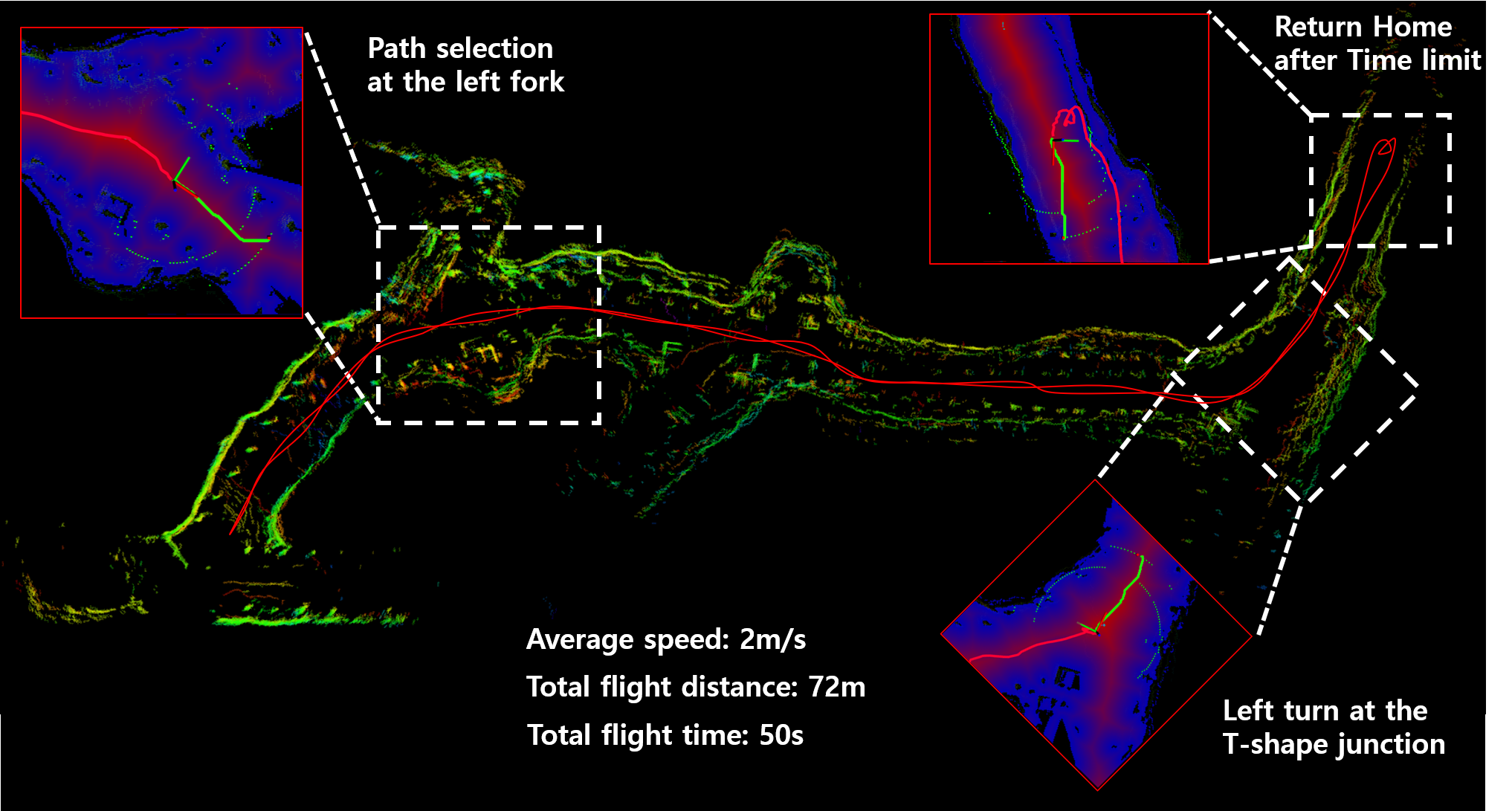}
    \caption{Autonomous exploration results at the Beckley coalmine. The drone explores long smooth curve while passing the left fork and discloses about 72 m for 50 s.\\ Location: Beckley mine, West Virginia, United States.}
    \label{fig:beckley_curve}
\end{figure}

\begin{figure}[t]
    \centering
    \includegraphics[width=3.2in]{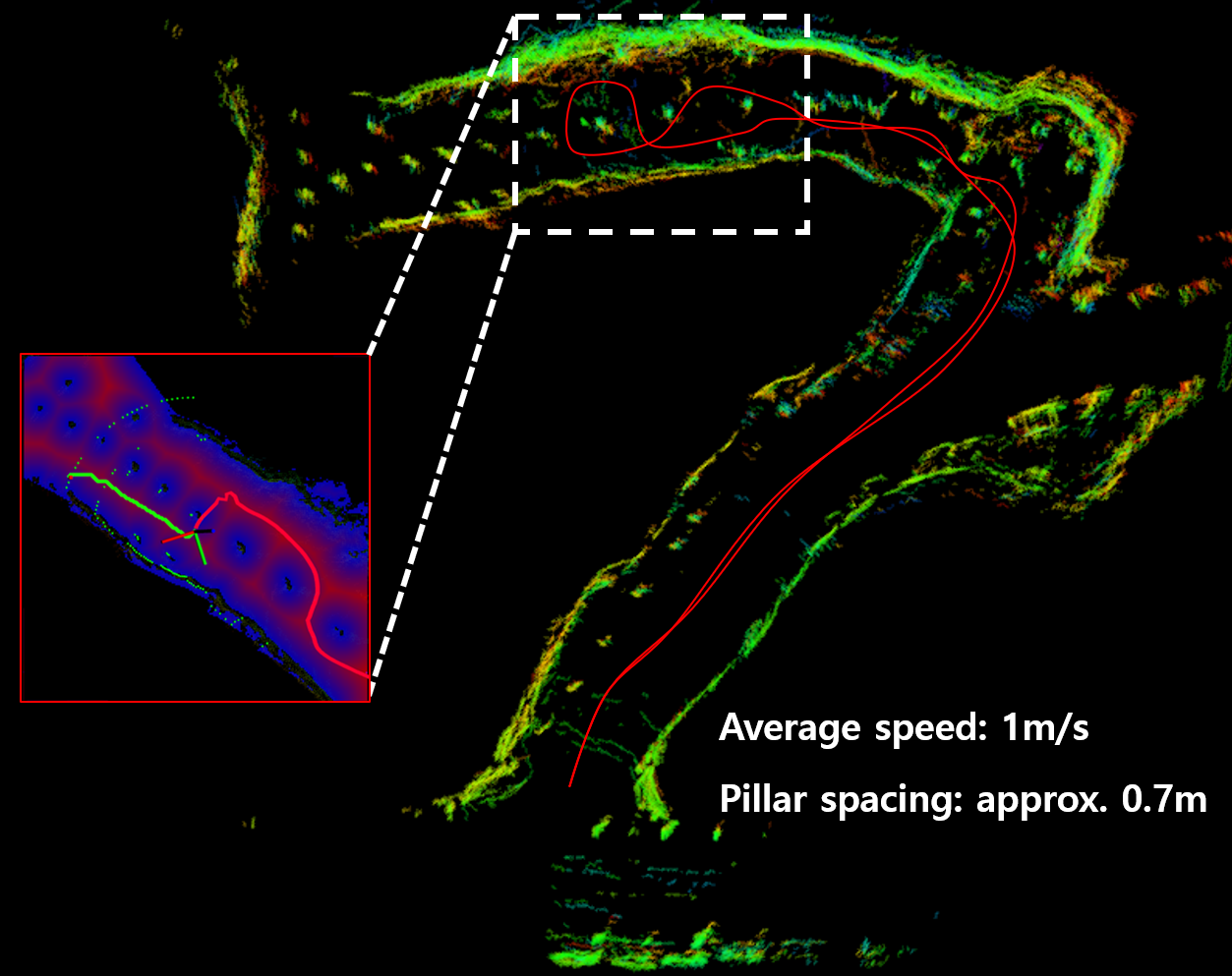}
    \caption{Autonomous exploration results at the Beckley coal-mine. Successful autonomous flight in a narrow pillar course \\ Location: Beckley mine, West Virginia, United States.}
    \label{fig:beckley_pillar}
\end{figure}

\begin{figure}[t]
    \centering
    \includegraphics[width=3.2in]{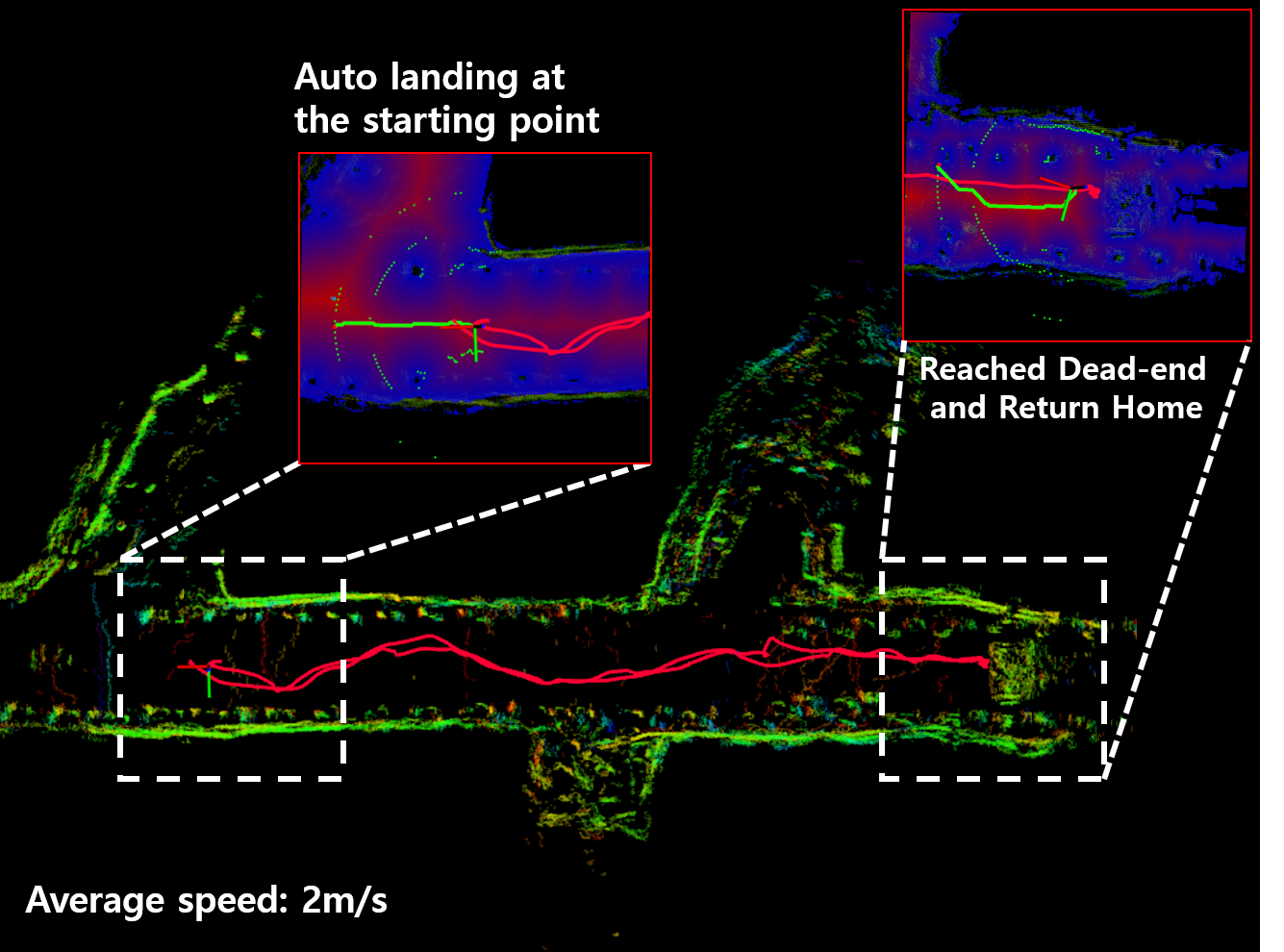}
    \caption{Autonomous exploration results at the Beckley coal-mine. Successful autonomous flight with turn-around action after detecting the dead-end\\ Location: Beckley mine, West Virginia, United States.}
    \label{fig:beckley_straight}
\end{figure}

The Beckley mine exploration, as an example, is presented in this section. The Beckley mine is a coalmine with a square-shaped structure and has about 3 m widths. Here, we tested three types of courses, which are curved, narrow corridors with pillars, and a straight path. The curved course has long curved corridors with a few junctions, and the narrow pillar course has a confined corridor with several pillars. Finally, the straight course is open to the left and the end of the course is blocked by obstacles.\\ 
\autoref{fig:beckley_curve} presents the curve type course test. The framework explores about 72 m for 50 s while passing the left fork and returning home due to the pre-set time limit (50 s). Narrow pillar example is shown in \autoref{fig:beckley_pillar}. This example shows the autonomy capability of the lightweight framework in confined and cluttered environment. The straight course example is presented in \autoref{fig:beckley_straight}. The drone reached the dead-end and returned home about 0.2 m apart from the starting point. The specification of the Beckley coal mine is presented in Table \ref{table:coalmineWV}.
\begin{table}[H]
\centering
\caption{Characteristic of the Beckley Mine}
\label{table:coalmineWV}
\begin{tabular}{|c|c|c|c|}\hline
\textit{\textbf{Parameter}} & \textit{\textbf{Value}}& \textit{\textbf{Parameter}}&\textit{\textbf{Value}} \\ \hline
Width & $\leq 3.0 m$& Name&Beckley coal mine \\ \hline 
Height & $\leq 1.5 m$& Location & West Virginia \\ \hline 
Tunnel Shape & Square shape   & Characteristic & Muddy \\ \hline
\end{tabular}
\end{table}

\section{Discussions}
The results provide a comprehensive evaluation of the performance of the proposed lightweight autonomy framework in various simulation and real-world environments. Each lightweight autonomy module, such as the local-map generation, the ESDF-map generation, end-point selection, and grid-search confer fully autonomous exploration capability to the real-world aerial robot. The end-point selection method selected a destination where the robot can safely navigate the local area. We demonstrated the successful exploration capability of the lightweight autonomy framework by fusing the end-point with the grid-search method, reflecting ESDF values.\\ 
A deeper consideration is that quadrotors mostly generate much dust when they fly in the mine. To address this, we used a software de-dust method by measuring the intensity value. However, we failed to detect thin objects and obstacles with less reflection in this case. Eventually, we solved the problem by replacing the hardware from Hokuyo UST-20LX, a passive laser scanner with RPLiDAR A3, an active laser scanner. However, more fundamental dust-rejection methods, such as multi-sensor fusion should be devised. 

\section{CONCLUSIONS}
\label{C_FW}
This study proposes a detailed autonomy framework for exploring unknown environments for a lightweight aerial robot. The proposed method utilizes the ESDF map to fly through a collision-free flight path, followed by an end-point selection method. The proposed lightweight autonomy framework has been experimented with various simulations and real-world environments, and successfully demonstrated the performance of the algorithm.\\
In the future, we extend our 2.5D mapping approach to 3D using the recently developed lightweight solid-state-LiDAR (\cite{l515}), to achieve full-scale planning and exploration for the citified areas, such as urban area, which requires a more agile vertical planning method. The method can be directly implemented to a 3D method where there exists a 3D occupancy grid map. In this way, the drone can fly a horizontal course and vertical passage, such as stairways. Finally, this paper considers mainly local planners and excludes information about global planners. By introducing of a hierarchical route planning method to effectively and meticulously observe the area under exploration, it will plan a global road map and conduct research on how to avoid obstacle collisions locally while moving along the road map.

%\addtolength{\textheight}{-12cm}
\bibliographystyle{IEEEtran} 
\bibliography{references}

\begin{thebibliography}{10}
\providecommand{\url}[1]{#1}
\csname url@samestyle\endcsname
\providecommand{\newblock}{\relax}
\providecommand{\bibinfo}[2]{#2}
\providecommand{\BIBentrySTDinterwordspacing}{\spaceskip=0pt\relax}
\providecommand{\BIBentryALTinterwordstretchfactor}{4}
\providecommand{\BIBentryALTinterwordspacing}{\spaceskip=\fontdimen2\font plus
\BIBentryALTinterwordstretchfactor\fontdimen3\font minus
  \fontdimen4\font\relax}
\providecommand{\BIBforeignlanguage}[2]{{%
\expandafter\ifx\csname l@#1\endcsname\relax
\typeout{** WARNING: IEEEtran.bst: No hyphenation pattern has been}%
\typeout{** loaded for the language `#1'. Using the pattern for}%
\typeout{** the default language instead.}%
\else
\language=\csname l@#1\endcsname
\fi
#2}}
\providecommand{\BIBdecl}{\relax}
\BIBdecl

\bibitem{balta2017integrated}
H.~Balta, J.~Bedkowski, S.~Govindaraj, K.~Majek, P.~Musialik, D.~Serrano,
  K.~Alexis, R.~Siegwart, and G.~De~Cubber, ``Integrated data management for a
  fleet of search-and-rescue robots,'' \emph{Journal of Field Robotics},
  vol.~34, no.~3, pp. 539--582, 2017.

\bibitem{bircher2016three}
A.~Bircher, M.~Kamel, K.~Alexis, M.~Burri, P.~Oettershagen, S.~Omari,
  T.~Mantel, and R.~Siegwart, ``Three-dimensional coverage path planning via
  viewpoint resampling and tour optimization for aerial robots,''
  \emph{Autonomous Robots}, vol.~40, no.~6, pp. 1059--1078, 2016.

\bibitem{milford2012seqslam}
M.~J. Milford and G.~F. Wyeth, ``Seqslam: Visual route-based navigation for
  sunny summer days and stormy winter nights,'' in \emph{2012 IEEE
  international conference on robotics and automation}.\hskip 1em plus 0.5em
  minus 0.4em\relax IEEE, 2012, pp. 1643--1649.

\bibitem{yap2009slam}
T.~N. Yap and C.~R. Shelton, ``Slam in large indoor environments with low-cost,
  noisy, and sparse sonars,'' in \emph{2009 IEEE International Conference on
  Robotics and Automation}.\hskip 1em plus 0.5em minus 0.4em\relax IEEE, 2009,
  pp. 1395--1401.

\bibitem{yamauchi1997frontier}
B.~Yamauchi, ``A frontier-based approach for autonomous exploration.'' in
  \emph{cira}, vol.~97, 1997, p. 146.

\bibitem{hart1968formal}
P.~E. Hart, N.~J. Nilsson, and B.~Raphael, ``A formal basis for the heuristic
  determination of minimum cost paths,'' \emph{IEEE transactions on Systems
  Science and Cybernetics}, vol.~4, no.~2, pp. 100--107, 1968.

\bibitem{connolly1985determination}
C.~Connolly, ``The determination of next best views,'' in \emph{Proceedings.
  1985 IEEE international conference on robotics and automation}, vol.~2.\hskip
  1em plus 0.5em minus 0.4em\relax IEEE, 1985, pp. 432--435.

\bibitem{kamal2005milp}
W.~Kamal, D.-W. Gu, and I.~Postlethwaite, ``Milp and its application in flight
  path planning,'' \emph{IFAC Proceedings Volumes}, vol.~38, no.~1, pp. 55--60,
  2005.

\bibitem{schouwenaars2004receding}
T.~Schouwenaars, J.~How, and E.~Feron, ``Receding horizon path planning with
  implicit safety guarantees,'' in \emph{Proceedings of the 2004 American
  control conference}, vol.~6.\hskip 1em plus 0.5em minus 0.4em\relax IEEE,
  2004, pp. 5576--5581.

\bibitem{kuwata2005robust}
Y.~Kuwata, T.~Schouwenaars, A.~Richards, and J.~How, ``Robust constrained
  receding horizon control for trajectory planning,'' in \emph{AIAA Guidance,
  Navigation, and Control Conference and Exhibit}, 2005, p. 6079.

\bibitem{shim2005autonomous}
D.~Shim, H.~Chung, H.~J. Kim, and S.~Sastry, ``Autonomous exploration in
  unknown urban environments for unmanned aerial vehicles,'' in \emph{AIAA
  Guidance, Navigation, and Control Conference and Exhibit}, 2005, p. 6478.

\bibitem{prazenica2006vision}
R.~Prazenica, A.~Kurdila, R.~Sharpley, and J.~Evers, ``Vision-based geometry
  estimation and receding horizon path planning for uavs operating in urban
  environments,'' in \emph{2006 American Control Conference}.\hskip 1em plus
  0.5em minus 0.4em\relax IEEE, 2006, pp. 6--pp.

\bibitem{bellingham2002receding}
J.~Bellingham, A.~Richards, and J.~P. How, ``Receding horizon control of
  autonomous aerial vehicles,'' in \emph{Proceedings of the 2002 American
  Control Conference (IEEE Cat. No. CH37301)}, vol.~5.\hskip 1em plus 0.5em
  minus 0.4em\relax IEEE, 2002, pp. 3741--3746.

\bibitem{bircher2016receding}
A.~Bircher, M.~Kamel, K.~Alexis, H.~Oleynikova, and R.~Siegwart, ``Receding
  horizon" next-best-view" planner for 3d exploration,'' in \emph{2016 IEEE
  international conference on robotics and automation (ICRA)}.\hskip 1em plus
  0.5em minus 0.4em\relax IEEE, 2016, pp. 1462--1468.

\bibitem{allen2016real}
R.~Allen and M.~Pavone, ``A real-time framework for kinodynamic planning with
  application to quadrotor obstacle avoidance,'' in \emph{AIAA Guidance,
  Navigation, and Control Conference}, 2016, p. 1374.

\bibitem{gao2017gradient}
F.~Gao, Y.~Lin, and S.~Shen, ``Gradient-based online safe trajectory generation
  for quadrotor flight in complex environments,'' in \emph{2017 IEEE/RSJ
  International Conference on Intelligent Robots and Systems (IROS)}.\hskip 1em
  plus 0.5em minus 0.4em\relax IEEE, 2017, pp. 3681--3688.

\bibitem{peng2012intelligent}
X.~Peng and D.~Xu, ``Intelligent online path planning for uavs in adversarial
  environments,'' \emph{International Journal of Advanced Robotic Systems},
  vol.~9, no.~1, p.~3, 2012.

\bibitem{lai2016robust}
S.~Lai, K.~Wang, H.~Qin, J.~Q. Cui, and B.~M. Chen, ``A robust online path
  planning approach in cluttered environments for micro rotorcraft drones,''
  \emph{Control Theory and Technology}, vol.~14, no.~1, pp. 83--96, 2016.

\bibitem{popovic2016online}
M.~Popovic, G.~Hitz, J.~Nieto, R.~Siegwart, and E.~Galceran, ``Online
  informative path planning for active classification on uavs,'' \emph{arXiv
  preprint arXiv:1606.08164}, 2016.

\bibitem{darpa_rss}
``https://www.darpa.mil/program/darpa-subterranean-challenge,'' [Online;
  accessed 08-Febuary-2021].

\bibitem{dang2020graph}
T.~Dang, M.~Tranzatto, S.~Khattak, F.~Mascarich, K.~Alexis, and M.~Hutter,
  ``Graph-based subterranean exploration path planning using aerial and legged
  robots,'' \emph{Journal of Field Robotics}, vol.~37, no.~8, pp. 1363--1388,
  2020.

\bibitem{dang2018visual}
T.~Dang, C.~Papachristos, and K.~Alexis, ``Visual saliency-aware receding
  horizon autonomous exploration with application to aerial robotics,'' in
  \emph{2018 IEEE International Conference on Robotics and Automation
  (ICRA)}.\hskip 1em plus 0.5em minus 0.4em\relax IEEE, 2018, pp. 2526--2533.

\bibitem{reinhart2020learning}
R.~Reinhart, T.~Dang, E.~Hand, C.~Papachristos, and K.~Alexis, ``Learning-based
  path planning for autonomous exploration of subterranean environments,'' in
  \emph{2020 IEEE International Conference on Robotics and Automation
  (ICRA)}.\hskip 1em plus 0.5em minus 0.4em\relax IEEE, 2020, pp. 1215--1221.

\bibitem{de2020collision}
P.~De~Petris, H.~Nguyen, T.~Dang, F.~Mascarich, and K.~Alexis,
  ``Collision-tolerant autonomous navigation through manhole-sized confined
  environments,'' in \emph{2020 IEEE International Symposium on Safety,
  Security, and Rescue Robotics (SSRR)}.\hskip 1em plus 0.5em minus 0.4em\relax
  IEEE, 2020, pp. 84--89.

\bibitem{tan2019design}
C.~H. Tan, D.~S. bin Shaiful, W.~J. Ang, S.~K.~H. Win, and S.~Foong, ``Design
  optimization of sparse sensing array for extended aerial robot navigation in
  deep hazardous tunnels,'' \emph{IEEE Robotics and Automation Letters},
  vol.~4, no.~2, pp. 862--869, 2019.

\bibitem{AuvideaJ120}
``https://auvidea.eu/j120/,'' [Online; accessed 08-Febuary-2021].

\bibitem{45466}
W.~Hess, D.~Kohler, H.~Rapp, and D.~Andor, ``Real-time loop closure in 2d lidar
  slam,'' in \emph{2016 IEEE International Conference on Robotics and
  Automation (ICRA)}, 2016, pp. 1271--1278.

\bibitem{oleynikova2016signed}
H.~Oleynikova, A.~Millane, Z.~Taylor, E.~Galceran, J.~Nieto, and R.~Siegwart,
  ``Signed distance fields: A natural representation for both mapping and
  planning,'' in \emph{RSS 2016 Workshop: Geometry and Beyond-Representations,
  Physics, and Scene Understanding for Robotics}.\hskip 1em plus 0.5em minus
  0.4em\relax University of Michigan, 2016.

\bibitem{pohl1973avoidance}
I.~Pohl, ``The avoidance of (relative) catastrophe, heuristic competence,
  genuine dynamic weighting and computational issues in heuristic problem
  solving,'' in \emph{Proceedings of the 3rd international joint conference on
  Artificial intelligence}.\hskip 1em plus 0.5em minus 0.4em\relax Morgan
  Kaufmann Publishers Inc., 1973, pp. 12--17.

\bibitem{ceres-solver}
S.~Agarwal, K.~Mierle, and Others, ``Ceres solver,''
  \url{http://ceres-solver.org}.

\bibitem{chrony}
``https://chrony.tuxfamily.org,'' [Online; accessed 08-Febuary-2021].

\bibitem{meier2011pixhawk}
L.~Meier, P.~Tanskanen, F.~Fraundorfer, and M.~Pollefeys, ``Pixhawk: A system
  for autonomous flight using onboard computer vision,'' in \emph{2011 IEEE
  International Conference on Robotics and Automation}.\hskip 1em plus 0.5em
  minus 0.4em\relax IEEE, 2011, pp. 2992--2997.

\bibitem{karaman2011anytime}
S.~Karaman, M.~R. Walter, A.~Perez, E.~Frazzoli, and S.~Teller, ``Anytime
  motion planning using the rrt,'' in \emph{2011 IEEE International Conference
  on Robotics and Automation}.\hskip 1em plus 0.5em minus 0.4em\relax IEEE,
  2011, pp. 1478--1483.

\bibitem{zhang2020falco}
J.~Zhang, C.~Hu, R.~G. Chadha, and S.~Singh, ``Falco: Fast likelihood-based
  collision avoidance with extension to human-guided navigation,''
  \emph{Journal of Field Robotics}, vol.~37, no.~8, pp. 1300--1313, 2020.

\bibitem{zhang2014loam}
J.~Zhang and S.~Singh, ``Loam: Lidar odometry and mapping in real-time.'' in
  \emph{Robotics: Science and Systems}, vol.~2, no.~9, 2014.

\bibitem{l515}
``https://www.intelrealsense.com/lidar-camera-l515/,'' [Online; accessed
  08-Febuary-2021].

\end{thebibliography}
\end{document}